\documentclass[conference]{IEEEtran}
\IEEEoverridecommandlockouts
\usepackage{tikz}
\usetikzlibrary{shapes,shadows,arrows}

\tikzstyle{decision} = [diamond, draw]
\tikzstyle{line} = [draw, -stealth, thick]
\tikzstyle{elli}=[draw, ellipse, minimum height=8mm, text width=5em, text centered]
\tikzstyle{block} = [draw, rectangle,  text width=8em, text centered,node distance=7em]
\usetikzlibrary{positioning}
\usepackage{pifont}
\usepackage{float}
\usepackage{booktabs}
\floatstyle{plain}
\newfloat{example}{thp}{lop}
\floatname{example}{\textsc{Example}}
\usepackage{xcolor}
\usepackage[export]{adjustbox}
\usepackage{cite}
\usepackage{amsmath,amssymb,amsfonts}
\DeclareMathOperator*{\argmin}{arg\,min}
\usepackage{bm}
\usepackage{algorithmic}
\usepackage{efbox,graphicx}
\usepackage{subfig}
\usepackage{textcomp}
\usepackage{xcolor}
\usepackage{pgfplots}
\usepackage{multirow}

\usepackage{url}
\usepackage{breakurl}
\usepackage[breaklinks]{hyperref}

\def\BibTeX{{\rm B\kern-.05em{\sc i\kern-.025em b}\kern-.08em
    T\kern-.1667em\lower.7ex\hbox{E}\kern-.125emX}}
    
\usepackage{amsthm}

\theoremstyle{definition}

\usetikzlibrary{matrix}

\begin{document}

\title{Linnaeus: A highly reusable and adaptable ML based log classification pipeline}

\author{\IEEEauthorblockN{Armin Catovic, Carolyn Cartwright, Yasmin Tesfaldet Gebreyesus and Simone Ferlin}\IEEEauthorblockA{Ericsson AB, Stockholm, Sweden\\Email: \{armin.catovic, carolyn.cartwright, yasmin.tesfaldet.gebreyesus, simone.ferlin\}@ericsson.com}}

\maketitle

\begin{abstract}
Logs are a common way to record detailed run-time information in software. As modern software systems evolve in scale and complexity, logs have become indispensable to understanding the internal states of the system. At the same time however, manually inspecting logs has become impractical. In recent times, there has been more emphasis on statistical and machine learning (ML) based methods for analyzing logs. While the results have shown promise, most of the literature focuses on algorithms and state-of-the-art (SOTA), while largely ignoring the practical aspects. In this paper we demonstrate our end-to-end log classification pipeline, Linnaeus. Besides showing the more traditional ML flow, we also demonstrate our solutions for adaptability and re-use, integration towards large scale software development processes, and how we cope with lack of labelled data. We hope Linnaeus can serve as a blueprint for, and inspire the integration of, various ML based solutions in other large scale industrial settings.
\end{abstract}


\section{Introduction}\label{section: introduction}
Over the past decade we have seen continued investments in machine learning (ML) and artificial intelligence (AI), across both industry and academia. These investments are not unwarranted - market analysts expect AI to potentially deliver additional economic output of around \$13 trillion USD by 2030, boosting global GDP by about 1.2 percent a year~\cite{mckinsey2018frontier}. Despite this optimistic outlook, there has been slow adoption of ML/AI in certain business areas; one reason is that the legacy systems and the ways of working are not prepared for integration with ML/AI \cite{mckinsey2018adoption}. At the same time, the overwhelming majority of research and publications focus on algorithms and state-of-the-art (SOTA), while the challenges of integrating ML/AI solutions into existing applications and ways of working remain largely unexplored.%One reason for this slow adoption is that the overwhelming majority of research and publications focus on algorithms and state-of-the-art (SOTA), while the challenges of integrating ML/AI solutions into existing applications and ways of working remain largely unexplored.

In their seminal paper, Sculley et al~\cite{sculley2015debt} argue that ML systems have a special capacity for incurring technical debt, because they have all of the maintenance problems of traditional code plus an additional set of ML-specific issues. In the last few years there has been more focus on ML/AI platforms and operations, and new sub-fields have emerged, including ModelOps, MLOps and AIOps. While this has been successful in placing more emphasis on data and model lifecycle management, a number of gaps related to industrial software processes still persist.

In this experience paper, we demonstrate our ML based log classification pipeline, "Linnaeus", named after the famous Swedish botanist Carl von Linn\'e. Linnaeus is a microservice/Docker based log classification pipeline that leverages supervised learning. Logs may consist of any text-based logs, including application logs, hardware logs, and continuous integration (CI) logs. The primary objective of our system is to map log snippets onto specific labels or categories, in order to reduce the time to find a fault (TTFaF) and improve root cause analysis (RCA).

Our aim in this paper is to demonstrate the many pieces required for a successful ML system integration within an industrial context. Linnaeus was from the start envisaged as a highly re-usable common component, with flexible deployment across both large enterprises as well as resource constrained embedded systems; it was also designed to work with different application logs with vastly different schema and semantics. Later, Linnaeus was augmented with "LinnDA", a log annotation tool, to help with the lack of high quality labeled data. Operation and maintenance (O\&M) and observability aspects were also an integral part of Linnaeus, allowing for ease of use and full visibility during both training and inference. The design choices described in this paper have enabled Linnaeus to be deployed across a dozen different projects in Ericsson, with minimal support. 

In summary, focusing on the overall system, the contribution of our paper is twofold: first, we outline end-to-end all the necessary components required for a successful log classification pipeline; second we describe various components and nuances required for a successful re-use and integration of such a pipeline in a large-scale industrial setting. The rest of this paper is structured as follows: section \ref{section: related_works} describes related works in this area; various components and design patterns are described in detail in section \ref{section: solution_overview}; section \ref{section: CI} provides an example of a successful Linnaeus integration in one of Ericsson's platform organizations; ethical considerations are mentioned in section \ref{section: ethical_considerations}; finally the key takeaways are summarized in section \ref{section: conclusion}.
\section{Related Work}\label{section: related_works}
%Logs are a common way to record detailed run-time information in software and are used as one of the main data sources for software development and maintenance. As modern software systems evolve into large scale and complex structures, logs have also become one type of fast-growing data in industry. In many cases, traditional ways of manually inspect logs became impractical. At the same time, applications increasingly demand online monitoring. However, getting good hold of this data amounts, giving it meaning, e.g., context, helps end users as well as software developers navigate through the complexity of these systems. With better analysis of the logs one can better visualize internal states of the system thus predicting and detecting anomalies or failures as well as helping to better decisions when it comes to log storage and retention. 
Inspired by research in the areas of software automated anomaly detection and software bug localization and reporting, we enlist the closest related works that helped in the development of Linnaeus.

In~\cite{li2013loganalysis}, the author experiments with different machine learning techniques and different test cases to automate log analysis to improve reliable testing of large scale systems. One of the claims is that manual log analysis is still a common practice and indispensable, however, it requires expert knowledge and is time consuming. In~\cite{thomas2013classifier}, the authors empirically investigate the effectiveness of a large space of classifier configurations and combinations for automated software bug localization. The main outcome of this work is to release the pressure on developers, especially when the code-base is large.
In~\cite{jonsson2012automatedanomaly}, the authors address the challenge of the manual, laborious, and inaccurate process of assigning anomaly reports to the correct teams of large software systems. Thus, the authors develop and validate a research prototype applying machine learning to automatically route real anomaly reports to the correct teams in a large organization. Similarly, in~\cite{Jonsson2016automatedbug}, the authors treat the challenge of software maintenance with automated software bug report assignment with machine learning, using a large-scale proprietary industry project as a case study, i.e., more than 50,000 bug reports from five development projects across the company. The goal is to use ML to assign the bug reports to the appropriate development teams.
In~\cite{jonsson2016automaticlocalization}, the authors address the challenge to reduce software bug turn-around time in large-scale software organizations. They propose a Bayesian approach, using classification to predict where bugs are located in components, i.e., a form of automated fault localization. In~\cite{nousiainen2009anomaly}, the closest related work to Linnaeus, the authors focus on the analysis of server log data with the goal of anomaly detection and anomaly prediction in a set of monitored servers. The authors use offline pre-recorded data with the final objective to design solutions to detect anomalies in real-time to prevent mission-critical servers from suffering a general service outage.

Finally, contrary to more recent works~\cite{he2018automatedlog, du2019onlinestreaming, zhu2019anomalydetection, liu2019logzip}, where the focus is limited to the challenges around log analytics, i.e., data ingestion, training and classification, Linnaeus goes one step further including all end-to-end system aspects, e.g., from data ingestion to O\&M integration interfaces and visualization in addition to training and classification. In the next section we describe Linnaeus in more detail.

\section{Solution Overview}\label{section: solution_overview}
Linnaeus is a microservice based log classification pipeline. Our intention with Linnaeus is to reduce the troubleshooting effort of large scale software systems. In 2017, software failures cost the US economy \$1.7 trillion USD, with most of this effort attributed to fixing errors and replicating issues \cite{raygun2017cost}. Our approach is to use supervised learning, or more specifically a classification task, to map a log snippet onto a specific category. This categorization could drastically reduce the problem space of a troubleshooter. Categories or labels can be defined in any manner appropriate for the troubleshooting task. For example, it can be as simple as binary classification (fault/no-fault), where the intent is to detect a fault. Alternatively it may be formulated as a multi-label classification task where the label corresponds to a fault category such as out-of-memory (OOM), network issue, or hardware failure. It is also possible to daisy chain Linnaeus instances, so we first perform fault detection, followed by fault categorization.

\begin{figure*}[h!]
    \begin{center}
        \includegraphics[width=0.95\textwidth]{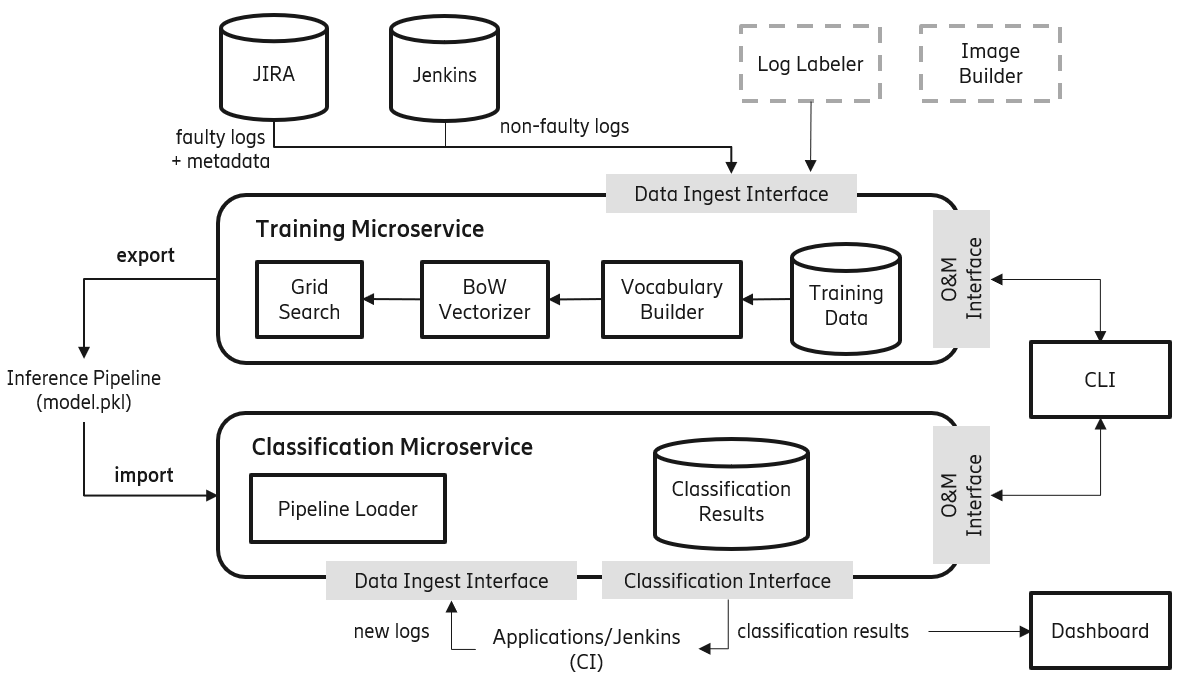}
    \end{center}
    \vspace{-4mm}
    \caption{High-level architecture of Linnaeus, the log classification pipeline presented in this paper.}
    \label{fig:linnaeus}
\end{figure*}

Fig. \ref{fig:linnaeus} shows the high-level architecture of Linnaeus, which is divided into two main components: training and classification. The training component loads the data, i.e. log snippets and their respective labels, and uses them to build a vocabulary and a bag of words (BoW) feature representation. A grid search method is used to find the best model and corresponding hyperparameters by exhaustively searching through the model/hyperparameter space. The result of training is a compressed classification pipeline, including the trained model, which can then be loaded and instantiated by the classification component. The classification component in turn performs the log categorization using new or unseen log data, stores the results in a time series database, and exposes the results over the classification interface. It is important to note that the entire process is completely autonomous. 

The inherent architecture, its various components, and the design patterns employed in Linnaeus accomplish several goals, namely they:

\begin{itemize}
    \item Allow rapid adaptation of the log classification pipeline across multiple and varied contexts without any code changes, i.e. within CI, on embedded telecom deployments, and across different log formats/semantics.
    \item Cater towards different ways of working across various product lines and organizations.
    \item Enable ease of use by abstracting the underlying infrastructure.
    \item Provide comprehensive observability of the underlying Linnaeus processes.
    \item Simplify the task of collecting high quality labeled data.
\end{itemize}
The subsequent sections expand on each component in more detail, as well as discuss additional support mechanisms such as O\&M and the log labeler/annotation tool.

%Figure~\ref{fig:linnaeus} shows the high-level architecture of Linnaeus, which is divided into two main steps: Training and classification. Logs (or text snippets) can be sent to training, where the data is prepared in a directory structure using grid search to find the best-fitting model. The result of the training step is a compressed classifier model, used as input in the classification step to classify faults based on the newly trained data.

%Linnaeus is based on open source implemented with \texttt{Python}. It also uses \texttt{Flask} for representational state transfer (REST) API handlers, \texttt{Gunicorn} as a REST-based HTTP server, \texttt{Scikit-Learn} as the machine learning framework, and \texttt{InfluxDB} as the lightweight time-series database to store the classification results. Next we describe each step of Linnaeus depicted in Figure~\ref{fig:linnaeus} in more detail.

\subsection{Data}
Since Linnaeus uses supervised learning, we require a dataset $D$ consisting of $N$ input/output pairs $D=\{(x_n, y_n)\}_{n=1}^N$, where $N$ is the sample size, $x$ is the input or more specifically, the feature vector representation of the text based log (see \textit{Feature Representation}), and $y$ is the target label or category. Labels generally represent some context or fault category, such that it helps the troubleshooter reduce the problem space. For example, $y \in Y$ where $Y = \{"OOM", "overload", "network", "hardware"\}$ would be a reasonable representation of fault categories.

The input/output pairs generally arrive from bug reporting systems such as JIRA \cite{jira2021}, and CI systems such as Jenkins \cite{jenkins2021}. The common practice in Ericsson is to use JIRA for providing faulty logs and their corresponding labels, while Jenkins is used for non-faulty logs. Given that data from these systems can be quite noisy, i.e. due to incorrect assignment of metadata and labels, we also provide a separate log labeler/annotation tool, which can be used by domain experts in more rigorous manner (see \textit{Log Labeler}). In practical terms, log/label pairs can be fed from any system, since we expose a simple representational state transfer (REST) interface. The REST interface allows for data to be sent to Linnaeus in JavaScript Object Notation (JSON) \cite{json2017} format. Linnaeus is quite flexible in terms of how this can be done. For example, log data can be provided as a string inside the JSON object, or it can be a unified resource identifier (URI) path to a local or remote log file. In the latter case, Linnaeus will fetch and process the remote file automatically.

It is important to note that in traditional ML literature, data is considered static or "frozen". In our system, both input data and target labels are continuously changing. This means that sample size $N$ and target set $y \in Y$ are changing with respect to the evolution of the software system and organizational ways of working. Linnaeus automatically retrains its model and pipeline with respect to these changes (see \textit{Training}).

\subsection{Feature Representation}
The input data (logs) are represented by word count vectors \cite{salton1975vector}. Since we assume independence between words, this is usually referred to as a bag-of-words (BoW) representation. More specifically, given a log file $j$, the corresponding feature vector will be $x_j = (w_{1,j}, w_{2,j}, ..., w_{P,j})$ where $P$ is the vocabulary length, or the total number of unique words in the log corpus. Note that "word" in our context can be any contiguous sequence of characters, usually referred to as n-grams. Linnaeus provides some flexibility in how logs are tokenized, so a word or a token may be something closer to an English language word, i.e. sequence of characters delimited by space or punctuation (e.g. "error"); it may also be a sequence of words (e.g. "kernel error"), or a sequence of characters of finite length $n$ (e.g. character n-gram "err", where $n = 3$). This latter representation may be desirable when a log file consists of inconsistent formatting and we wish to avoid explicit parsing and pre-processing, e.g. JSON format may be interleaved with syslog \cite{syslog2009}.

Linnaeus further allows the option of scaling word count vectors using a method called term frequency-inverse document frequency (TF-IDF) \cite{salton1987weighting}. Given a word $w_p$ and a log file $x_j$, we multiply its term frequency $tf(w_p, x_j)$ by inverse document frequency $idf(w_p, X) = \log {\frac{N}{1 + \lvert \{ x \in X : w_p \in x \} \rvert}}$, where $N$ represents the size of sample space, or all the log files in the log corpus $X$, while the denominator in the $idf$ expression represents the number of log files where the word $w_p$ appears; the expression $1 + ...$ in the denominator prevents division-by-zero errors. TF-IDF can be interpreted as penalizing words that occur very frequently in the entire log corpus and provide very little discriminative power as a result.

As mentioned previously, the data presented to Linnaeus will change over time as the software evolves. This means that as the vocabulary and its size $P$ change, Linnaeus will automatically adopt these changes.

\subsection{Training}
% Armin's Comments:
%   - Provide mathematical description of the classification process, i.e. f: X->Y where X are features and Y are labels; data is presented as set of tuples {(x1, y1), (x2, y2), ..., (xn, yn)}, we aim to minimize loss function, i.e. argmin...
%   - Provide citation for TfIdf (e.g. Karen Jones paper)
%   - Mention .pkl more explicitly as Python "pickle"; important to state that we prepare our architecture in such way that it can be pickled - this is important because you can't use lambda functions during pickle (there is a stackoverflow reference on that!
%   - Mention that we actually do train/test split internally and that grid search works against the test set
%   - Mention that by default we provide a number of different algorithms; linear algorithms like logistic regression perform just as well as Gaussian Naive Bayes, and SVMs.
The task of the training component is to learn a classifier $f(x; \theta)$ such that $f$ provides a mapping from inputs (log files or snippets) $x \in X$, to outputs (categories) $y \in Y$, i.e. $f : X \rightarrow Y$; $\theta$ are the model parameters we wish to learn. The parameters can be learnt through empirical risk minimization $\hat{\theta} = \argmin_{\theta} L(\theta) = \argmin_{\theta}\frac{1}{N}\sum_{n=1}^N l(y_n, f(x_n; \theta))$, where $l$ represents a risk or a loss function, i.e. the difference between predicted label $\hat{y}$ and the actual category $y$.

Internally, our training component implements the \texttt{TrainWorker} class, which wraps Scikit-Learn's \texttt{Pipeline} module \cite{sklearn2021pipeline}. After receiving the data over the REST interface, the \texttt{TrainWorker} performs a train/test partition, before passing the data through the ML pipeline. The pipeline includes the tokenizer, the vocabulary builder, vectorizer, i.e. feature representation, and finally the grid search function, which performs the model training and hyperparameter tuning. Linnaeus can automatically choose one of a number of learning algorithms, including Gaussian Naive Bayes, Support Vector Machine (SVM), Logistic Regression, and Random Forest. Compared to more complex deep learning approaches, our choice of "classical" ML algorithms provides two main advantages: first, it gives better interpretability while achieving good accuracy, i.e. we can easily identify which words are representative in each category; second, training is faster and consumes less memory. In addition, LSTMs and auto-encoders performed worse than the "classical" ML algorithms. The output of the training component is the learned model as well as the actual pipeline just described. The pipeline needs to be preserved in its exact form since Linnaeus needs to understand how to tokenize and vectorize the new log data. The model and the pipeline are neatly packaged using Python's \texttt{pickle} module \cite{python2021pickle}, and can subsequently be loaded and instantiated by the classification component (see \textit{Classification}).

Flexibility is one of the key attributes of the Linnaeus architecture. While the main flow of the training microservice remains the same in every deployment or situation, Linnaeus is highly configurable so that it can be reused in different deployments or situations by simply tweaking certain configuration parameters. A good example of where this can be necessary and where Linnaeus allows changes through configuration is with stop words. Stop words are removed from a model’s vocabulary to reduce noise and false positives that they cause because they occur frequently but are meaningless in the problem context. However, stop words can be meaningless in one scenario while very important in another depending on the data or the problem being analysed. The Linnaeus architecture allows the configuration of the training microservice and the Docker container it runs on through its command line interface (CLI) parameters, as well as a configuration file. It can be configured to:
% Very good point on stopwords! Stopwords list will have different impact in different organizations, depending on their WoW; for example in some cases "error" will be very important, while in other cases "error" should actually be ignored!
\begin{itemize}
    \item Tokenize on a word or character basis and specify the number of words or characters there are in the sequence.
    \item Remove a specified set of stop words from the model’s vocabulary.
    \item Set the minimum and maximum document frequency where those words that are below the minimum document frequency and those above the maximum frequency are filtered out of the vocabulary.
    \item Enable/disable feature vector scaling, i.e. using TF-IDF.
    \item Enable regularization to avoid overfitting, i.e. using L1 or L2 regularization.
    \item Take advantage of multiprocessing environments by creating a specified number of parallel jobs.
    \item Retrain the model at regular, specified time intervals, e.g. once a day.
    \item Use a specified set of training labels.
    \item Export a model to a specified local or remote directory or shared volume.
    \item Train using a specified learning algorithm.
\end{itemize}

\subsection{Classification}
% Armin's Comments:
%   - Mention that classifier loads the pickled pipeline and trained model and instantiates them both; the rest of the classifier code is just the API that talks to the classification pipeline and exposes results over REST interface; plus we have the results database (timeseries db)
The classification microservice is designed to run completely independently from the training microservice. This allows the training component to be deployed in one location (e.g. Ericsson R\&D premises), while the classification component is deployed on another location, such as the communications service providers' (CSPs) network equipment. New models and ML pipelines can be pushed out from the training component to CSP's premises as-needed (e.g. when significant model updates are made), or on regular intervals. This separation allows us to apply special build patterns on the classification component in order to keep its memory footprint as low as possible (see \textit{Image Builder}).

Internally, the classification component implements the \texttt{ClassifyWorker} class, which simply loads and instantiates the previously pickled model and pipeline. The pipeline receives new log data over the REST interface, and performs tokenization and feature vectorization before passing the feature vector through the trained model. The output of the model, i.e. the predicted label/category, is stored in a time-series database, and the classification result is sent in the response message back to the user. The classification interface allows classification on single logs, or log bundles consisting of many logs. In addition, it allows the user to provide a URI where the classification microservice can fetch the data, in case the data is stored remotely. 

%The Linnaeus architecture allows the configuration of the classification microservice and the Docker container it runs on through its CLI start command as well as a configuration file. It can be configured to:
Similar to the training component, the classification component can be configured via CLI parameters or a configuration file. It can be configured to:

\begin{itemize}
    \item Store only the classification results where the predicted category has a confidence score above a specified threshold.
    \item Specify a window/buffer in terms of the number of log lines to be considered for classification at the same time; this is especially useful when dealing with very large log files.
    %\item Ensure the relevant logs are classified together by indicating the number of log lines to classify I per log snippet.
    %\item Store only the classification results for a specified category that are above a specified value.
\end{itemize}

%\subsection{Data Ingestion Interface}
%Xxxx
% Armin's Comments:
%   - Mention that it's either using shared volumes or over REST API
%   - REST API uses JSON encoding, and more specifically we follow JSend format: https://github.com/omniti-labs/jsend
%   - For REST API we have lot of flexibility: sending log data as a raw string, sending log data as JSON encoding, or sending a path or URI to the log file, or even log bundle (e.g. zipped)

\subsection{Operation and Maintenance}
Monitoring an ML system is important to ensure it continues to operate effectively and that the quality of its predictions remain at the level that it was upon deployment. It is also tricky due to the CACE principle: Change Anything Changes Everything~\cite{sculley2015debt} and the inevitability that things change in a dynamic environment. 
\par The Linnaeus architecture uses several strategies to mitigate the effects of changes on the model’s performance:
\begin{itemize}
    \item By separating the training and classification microservices, Linnaeus allows for automated, periodic retraining of the model without affecting the deployed classifier. Retraining captures changes over time in the relationships between input and output data that may well be served better by different hyperparameters or even algorithms. When a new model is produced that provides better metrics than the old model, the new model can be exported by the training service and imported for immediate use by the classifier without any downtime.
 \item Linnaeus uses asynchronous communication via a REST interface with JSON encoded messages. 
 		  \begin{itemize}
        \item For the training service, the REST application programming interfaces (APIs) allow changes to the training data. As the model input data characteristics and distribution drift away from their original characteristics and distribution, these APIs can be used to seamlessly introduce new categories and new training data points that will be picked up on the next round of automated, periodic retraining.
        \item For the classification service, the REST APIs allow for the export of the classification database, that is, a series of classifications over time including the text classified and the resulting classification. It allows a human to visually check to ensure the classifier is working as expected, providing transparency and explainability of an application’s actions based on the predictions.
        \item For both the training and classification microservices, the REST APIs allow a user to obtain and compare metrics. A comparison between the training service metrics and the classification service metrics can be performed to determine if a new trained model should be deployed. A comparison between the metrics that the classification microservice produced on the day of deployment and the metrics it currently produces can be performed to detect degradation of the quality of the classifier’s predictions over time.
    \end{itemize}
    \item The Linnaeus logging system produces logs showing the evaluation metrics of the trained model as well as the training time and latency of the prediction.
\end{itemize}

Life cycle management of Docker containers can be a source of frustration for the user, as Docker nomenclature and command syntax can be quite complex. To ease this complexity and allow rapid deployment, Linnaeus provides a simple wrapper CLI interface with the commands \texttt{install}, \texttt{remove}, \texttt{start}, \texttt{status} and \texttt{stop}. The \texttt{install} command in particular is flexible enough to either install a Docker image from a local directory, or from a common Docker repository. This can be useful when a specific version of the data and Docker image is needed.
%\par Only the \texttt{start} command has parameters beyond selecting which container type, train or classify, is being managed. All other details are addressed under the hood. The start CLI command is slightly more complex as the model may require some additional configuration parameters or the use of a shared directory or a port. Even so, placement of these parameters in the Docker command syntax is completely hidden from the user.
%\par Furthermore, the \texttt{install} CLI command is flexible enough to either install a Docker image from a directory or pull it and its data from a common Docker repository before installation. This can be useful when you need to pull a specific version of the data and Docker image.

\subsection{Image Builder}
One of the aims of Linnaeus is to enable the deployment of the classification component even on resource constrained systems. For example, some of the lower-end telecom hardware is limited to 2 GiB RAM. Since the Linnaeus classifier is packaged as a Docker container, our goal is to keep the container image size as small as possible. One way we accomplish this is by using Python's \texttt{trace} module \cite{python2021trace}. During the build process, we first run the classifier through a synthetic dataset, while \texttt{trace} collects statement and module coverage listings. Here it is important that the classifier's interfaces and functionality are fully exercised. The results from \texttt{trace} are then used to remove all Python modules that are not present in the output listings. We then apply further, more generic removal actions:

\begin{itemize}
    \item Removal of package manager caches, usually stored in \texttt{/root/.cache/}.
    \item Removal of Python's \texttt{Wheel} and \texttt{Egg} package data, i.e. \texttt{dist-info} and \texttt{egg-info} directories.
    \item Removal of unused binaries and shared libraries - these are specified in advance.
    \item Removal of symbol tables from the remaining binaries and shared libraries, i.e. using *nix \texttt{strip -s} command.
\end{itemize}

All of the above are provided as statements in a \texttt{Dockerfile}, which is used during the image build process. From our experience, we have succeeded in reducing the final classifier image size from 2 GiB to 150 MiB. A side-effect from our image builder is that the classifier component is also telecom-hardened, meaning that its vulnerability/attack surface has been significantly reduced by removing unneeded modules. This is important to consider in any critical infrastructure setting, since modern Python based ML frameworks consist of large number of 3rd party components.

\subsection{Log Labeler}
\begin{figure}[h!]
    \begin{center}
        \includegraphics[width=0.475\textwidth]{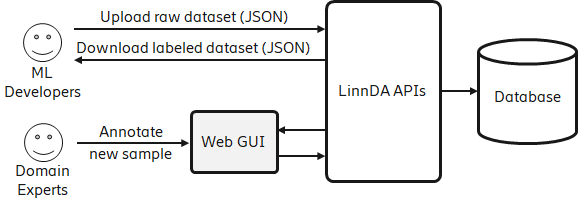}
    \end{center}
    \vspace{-4mm}
    \caption{Linnaeus data annotation tool (LinnDA) high-level architecture.}
    \label{fig:linnda}
\end{figure}

\begin{figure}[h!]
    \begin{center}
        \includegraphics[width=0.5\textwidth]{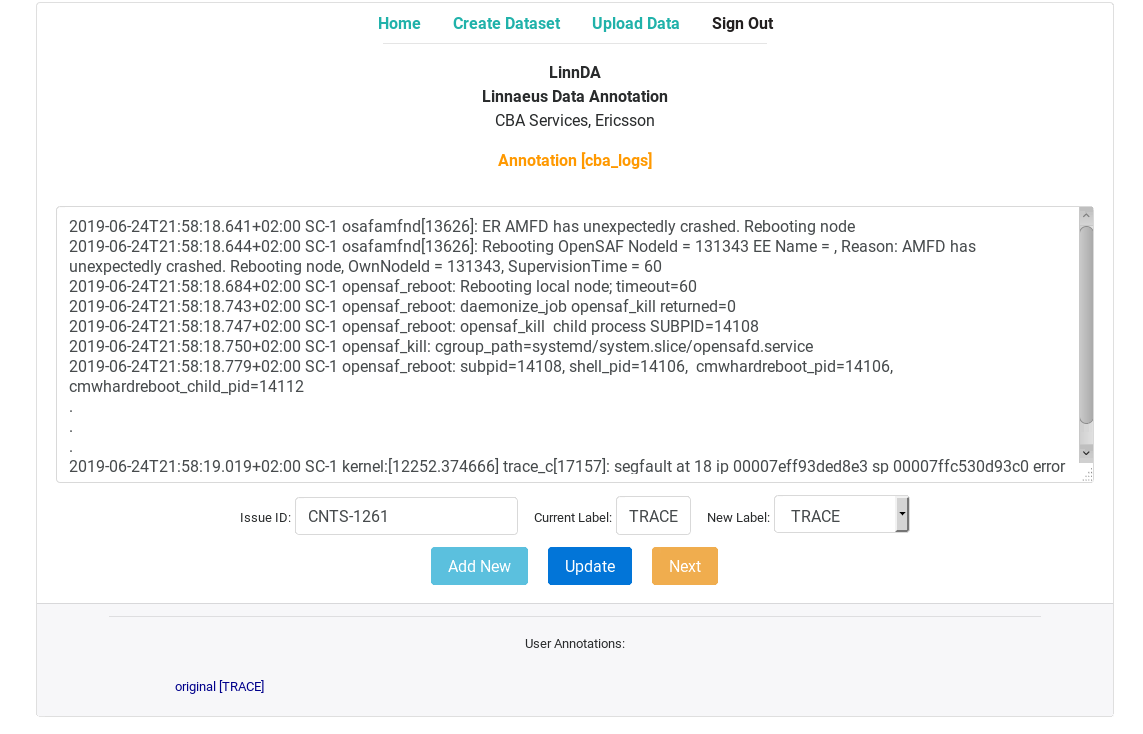}
    \end{center}
    \vspace{-4mm}
    \caption{LinnDA GUI used during the log annotation process.}
    \label{fig:linnda_gui}
\end{figure}

\begin{figure}[h!]
    \begin{center}
        \includegraphics[width=0.5\textwidth]{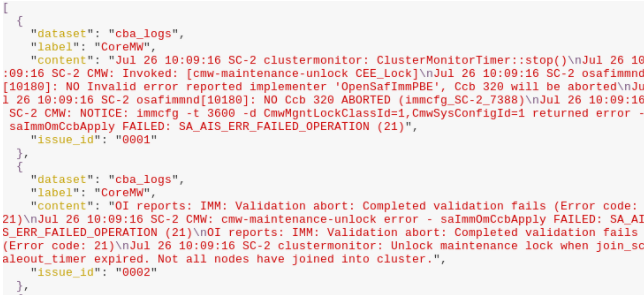}
    \end{center}
    \vspace{-4mm}
    \caption{Example of the annotated data samples produced by the LinnDA tool.}
    \label{fig:json_training_data}
\end{figure}

LinnDA (Linnaeus data annotation) is a log annotation service for facilitating the creation of clean datasets required for training the Linnaeus models. This service is used during the data acquisition stage of the machine learning process. LinnDA is designed to have all the generic features expected in a text annotation service (see Fig. \ref{fig:linnda} for the high-level architecture). LinnDA has a user friendly interface (see Fig. \ref{fig:linnda_gui}) in which domain experts, i.e. developers and troubleshooters, can create, upload, delete and export datasets along with their annotations/labels.  Annotation history is recorded and displayed to create complete transparency during the verification of datasets. For data exchange, Linnaeus and LinnDA use JSON format, where the JIRA id, component name, label, and the actual log snippet are included as individual fields (see Fig. \ref{fig:json_training_data}).

\section{Case Study: Deploying Linnaeus in Ericsson Continuous Integration Pipeline}\label{section: CI}
The following section describes how Linnaeus was successfully deployed within Ericsson's Component Based Architecture (CBA) platform organization and how it works inside CBA's nightly CI pipeline. Since CBA is composed of over a dozen different software components, each with their own development teams, our aim is to automatically categorize faulty logs according to the component or team responsible. The primary objective is to assign the troubleshooting task to the correct CBA team.

The nightly CI pipeline runs both functional and non-functional test cases and produces logs that are fed to Linnaeus for classification. The test cases for this pipeline are written using a Java based test automation framework, JCAT. When a failure is detected during test execution, system under test (SUT) logs are collected from the time of failure and back, by a configurable amount of time (default is 5 seconds), and then sent to Linnaeus for classification. Linnaeus returns the predicted label and corresponding confidence score, which are then added to a web based report generated as a result of the test. A link to the Linnaeus dashboard/GUI, run as a separate web service, is also included in the report and shows more details from the classification task. There, users can check the interpretation of the different sections of the log displayed along with the predicted accuracy of each component used during classification. Additionally, users have the option to request re-classification of logs directly from the Linnaeus GUI in order to get an updated result. This is useful in cases where Linnaeus has been trained with a new dataset and has a model with improved performance. 

\begin{figure*}[h!]
    \begin{center}
        \includegraphics[width=0.95\textwidth]{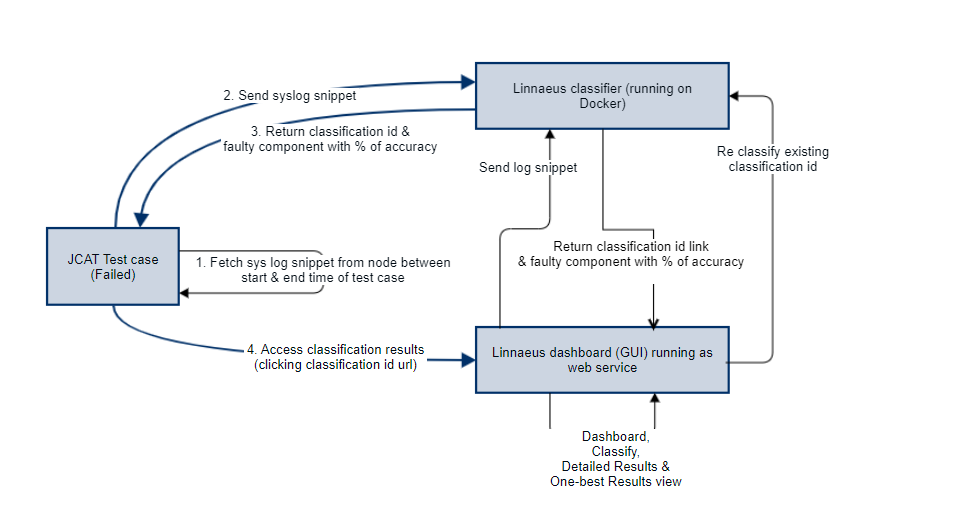}
    \end{center}
    \vspace{-8mm}
    \caption{Linnaeus continuous integration architecture.}
    \label{fig:jcat_integration}
\end{figure*}

Limited storage for logs was one of the main challenges while integrating Linnaeus. To solve this issue the default lightweight database was replaced with a more powerful enterprise one. This was made possible by the explicit separation of concerns inherent in the Linnaeus design. This flexibility in design has enabled CBA's developers and troubleshooters to benefit from continuous model performance improvements due to the increase in quality and quantity of the training data. Fig. \ref{fig:jcat_integration} shows the overall continuous integration architecture.

\begin{figure}[h!]
    \begin{center}
        \includegraphics[width=0.5\textwidth]{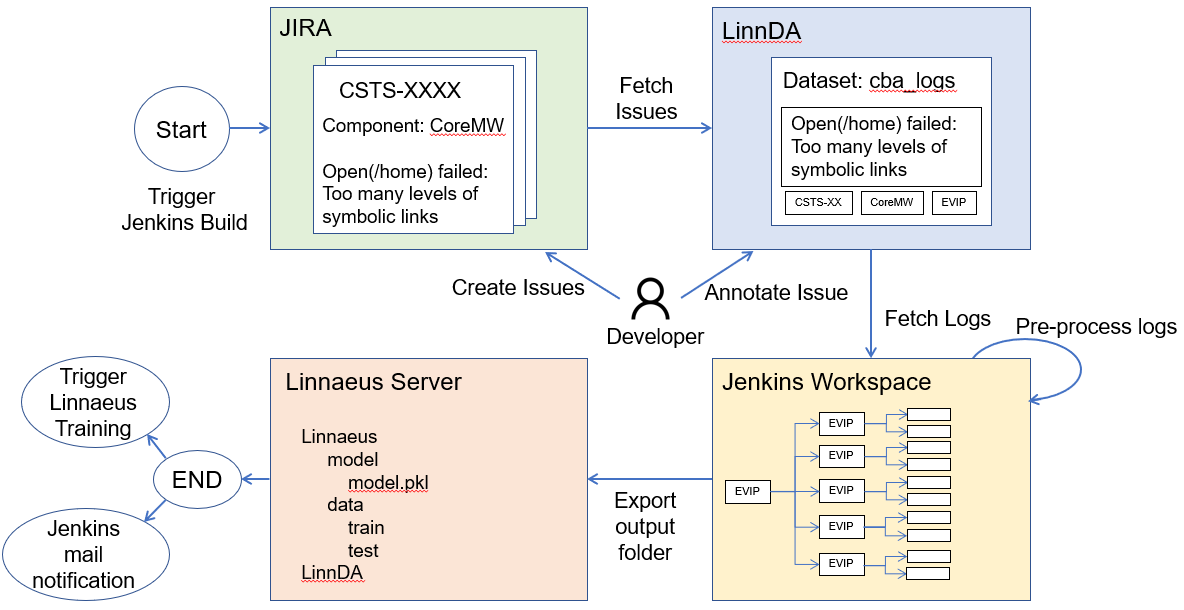}
    \end{center}
    \vspace{-4mm}
    \caption{Automated training data preparation pipeline.}
    \label{fig:auto_training_data}
\end{figure}

\begin{figure}[h!]
    \begin{center}
        \includegraphics[width=0.5\textwidth]{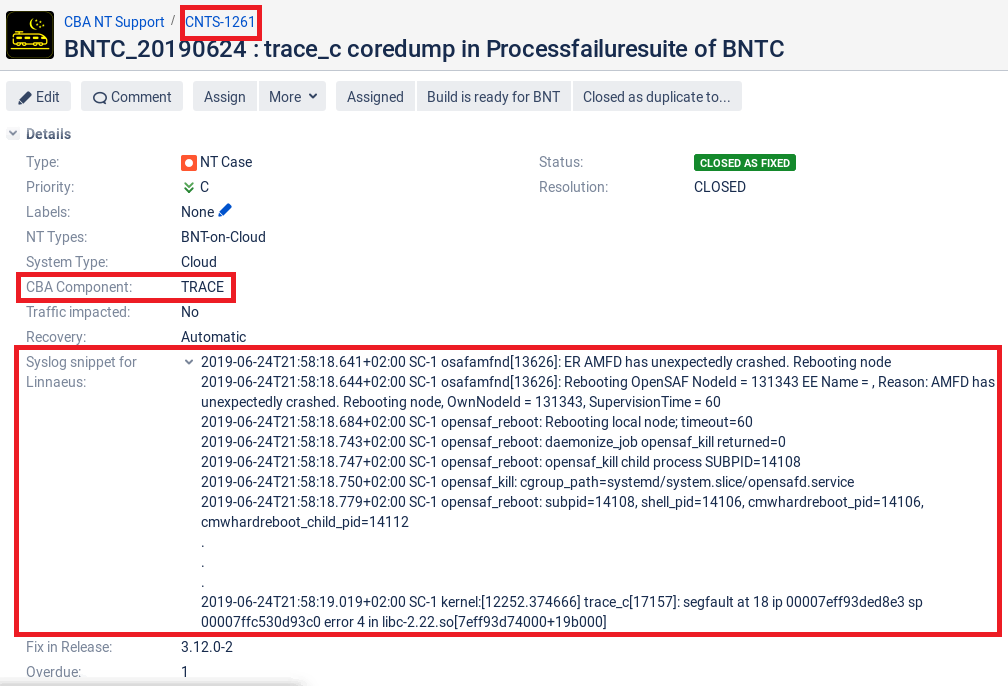}
    \end{center}
    \vspace{-4mm}
    \caption{JIRA ticket for collecting the input from users.}
    \label{fig:jira_TR}
\end{figure}

Another important feature of Linnaeus is the ability to automatically trigger re-training based on new data accumulated from the continuous integration pipeline. As shown in Fig. \ref{fig:auto_training_data}, the source of raw training data is a collection of JIRA trouble report (TR) tickets. Fig. \ref{fig:jira_TR} is a sample TR ticket. A weekly Jenkins job is triggered to collect logs and their respective labels from JIRA TR's closed on the previous week. This data is then exported to LinnDA (Linnaeus data annotation tool) and is stored there for a week so that domain experts can have user friendly access to verify the provided data annotation and do some data cleaning. Another weekly Jenkins job exports the verified dataset from LinnDA to Linnaeus and triggers the model training using the new training dataset. Once the model completes re-training, a model performance evaluation report is sent to the development team who then decides if a new classification model should be used.

\section{Ethical Considerations}\label{section: ethical_considerations}
Since Linnaeus focuses on automating a labour-intensive task it is natural to assume it poses a threat in terms of eliminating human jobs. Our stance is that this is not the case. As mentioned in section \ref{section: solution_overview}, software failures are a major dent to the economy, while the average software developer spends 32 hours each month fixing errors and replicating issues \cite{raygun2017cost}. Our informal discussions with software development teams in Ericsson mirror our own experiences, namely that it is in fact desirable to reduce tedious log analysis tasks. Developers can then spend more time on actual value creation.

Another potential ethical concern is regarding privacy and abuse of user-sensitive information. This however is not specific to Linnaeus nor our work presented here. Internal software and hardware logs should never expose user-sensitive data. Doing so could jeopardize user integrity whether we use Linnaeus or not. We could argue that Linnaeus actually helps preserve user privacy since it hides the underlying log data and simply outputs desired labels/categories. Having said that, if sensitive data is present in software logs and there is malicious intent on the part of a Linnaeus user, it would cause ethical concerns.
%Any user permission, any user data, any PII or IPR-related data or any "disturbance" on third-party systems that could be disclosed in the logs? Formulate.
\section{Discussion and Conclusion}\label{section: conclusion}
While there has been significant effort over the past decade developing data driven methods for automated log analysis and troubleshooting, the literature has focused exclusively on algorithms and methods, and their performance. At the other end of the scale, recent developments in MLOps and AIOps have put more emphasis on managing data and models, along with the underlying infrastructure and platforms. However, the various pieces and nuances required for successful integration towards products, applications and the ways of working, have largely been missed.

In this paper we have demonstrated, through our own experience in building and deploying our log classification pipeline Linnaeus, the various components and design factors that need to be considered on industrial scale. These extend far and beyond just algorithms and platforms. We can summarize our key takeaways as follows:

\begin{itemize}
    \item Large technology organizations with multiple product offerings aim to leverage software re-use as much as possible in order to improve their bottom line. In our case we enable re-use by:
    \begin{itemize}
        \item Packaging Linnaeus such that it can be run and controlled without any code changes and without the need to understand the underlying IT infrastructure.
        \item Placing emphasis on low memory footprint so that Linnaeus can also be deployed on resource constrained systems, which are commonplace in telecom. We achieve this using special build patterns as described in section \ref{section: solution_overview}. Our method implicitly takes care of some security concerns as well by minimizing the attack surface, i.e. we only use the modules that we need and nothing else.
        \item Providing a very flexible REST interface that caters to nearly all ways of working across various Ericsson design organizations.
        \item Implementing enough flexibility in terms of feature representation and model selection such that Linnaeus classification performance is optimal irrespective of the underlying log format and semantics. For example, using character n-grams Linnaeus can work well with both syslog and JSON log formats, i.e. no specialized pre-processing is required if both formats are present in the same log.
    \end{itemize}
    \item Supervised learning requires high quality labelled data. In Linnaeus deployments, these data and corresponding labels typically come from internal bug tracking systems such as JIRA, and CI systems such as Jenkins. However, ways of working vary from one team to the next, so such data can be noisy and in limited supply. We therefore provide an easy to use log annotation tool, LinnDa, that allows domain experts to automatically annotate and feed data into the Linnaeus pipeline.
    \item Telecommunications are considered critical infrastructure and therefore significant emphasis is placed on observability mechanisms that help trace and understand system behaviour. We have adopted that same approach with Linnaeus, by implementing extensive logging and observability within Linnaeus itself.
\end{itemize}

Going forward, we see additional improvements. The labelling tool can be further improved by applying weak supervision and data programming methods, allowing us to amass greater volume of labeled data. Adding explainability modules such as SHapley Additive exPlanations (SHAP) can help troubleshooters and data scientists alike better understand the reasons for the classification results. Finally, we can explore further footprint reductions and the application of TinyML, which is highly relevant for IoT and severely resource constrained devices.
%Some comments - elaborate, formulate:
%- It would be useful to provide a user guide on how to interpret the data.
%- Given a set of models, Linnaeus can be enhanced to find the best model to fit the data automatically when it creates the training pipeline and/or when it retrains the model.
%- The evaluation data coming from the REST API needs post processing to become easily readable.
%- Include a more complete list of machine learning algorithms to use so a product can choose one that best fits the unique characteristics of its dataset. Include instructions on when to use them and why so developers can make informed decisions.
%- Logs that are labeled as "fault" or "no-fault" do not give enough information to Linnaeus for it to classify the information since Linneaus is designed for supervised learning. For these cases, either a way to further classify the data in Linnaeus or modification to Linnaeus to handle unsupervised learning is needed.
%- Further testing of the speed of the vocabulary builder and the tokenizer is needed to ensure that they are faster than the scikit-learn built-in versions.
%- Change out the underlying machine learning model in the production environment via a REST API.

\bibliographystyle{IEEEtran}
\bibliography{references}

\end{document}